\documentclass[10pt,twocolumn,letterpaper]{article}
\usepackage[accsupp]{axessibility}
\usepackage{iccv}
\usepackage{times}
\usepackage{epsfig}
\usepackage{graphicx}
\usepackage{amsmath}
\usepackage{amssymb}

\usepackage{multirow}
\usepackage{multicol}
\usepackage{booktabs}
\usepackage{array}
\usepackage{threeparttable}
\usepackage{physics}
\usepackage{subfig}
\usepackage{bm}
\usepackage{dblfloatfix}
\usepackage{newtxmath}
\usepackage{dsfont}
\usepackage{color}
\usepackage{xcolor}
\usepackage{tabu}
\graphicspath{ {./images/} }

\usepackage[pagebackref=true,breaklinks=true,letterpaper=true,colorlinks,bookmarks=false]{hyperref}

 \iccvfinalcopy 

\def\iccvPaperID{3989} 

\ificcvfinal\fi

\begin{document}

\title{Spatial and Semantic Consistency Regularizations for Pedestrian Attribute Recognition}

\author{Jian Jia$^{\,1,2}$, Xiaotang Chen$^{\,1,2}$, Kaiqi Huang$^{\,1,2,3}$ \thanks{Corresponding author.}\\
	$^{1}$ the School of Artificial Intelligence, University of Chinese Academy of Sciences\\
	$^{2}$ CRISE, Institute of Automation, Chinese Academy of Sciences\\ 
	$^{3}$ CAS Center for Excellence in Brain Science and Intelligence Technology\\
	{\tt\small jiajian2018@ia.ac.cn, \{xtchen,kqhuang\} @nlpr.ia.ac.cn}
}

\maketitle

\ificcvfinal\thispagestyle{empty}\fi

\begin{abstract}
  While recent studies on pedestrian attribute recognition have shown remarkable progress in leveraging complicated networks and attention mechanisms, most of them neglect the inter-image relations and an important prior: spatial consistency and semantic consistency of attributes under surveillance scenarios. The spatial locations of the same attribute should be consistent between different pedestrian images, \eg, the ``hat" attribute and the ``boots" attribute are always located at the top and bottom of the picture respectively. In addition, the inherent semantic feature of the ``hat" attribute should be consistent, whether it is a baseball cap, beret, or helmet. To fully exploit inter-image relations and aggregate human prior in the model learning process, we construct a Spatial and Semantic Consistency (SSC) framework that consists of two complementary regularizations to achieve spatial and semantic consistency for each attribute. Specifically, we first propose a spatial consistency regularization to focus on reliable and stable attribute-related regions. Based on the precise attribute locations, we further propose a semantic consistency regularization to extract intrinsic and discriminative semantic features. We conduct extensive experiments on popular benchmarks including PA100K, RAP, and PETA. Results show that the proposed method performs favorably against state-of-the-art methods without increasing parameters. \end{abstract}
  
\vspace{-0.5em}

\section{Introduction}

\begin{figure*}
	\centering
	\subfloat[Spatial consistency on the ``short sleeve", ``boots", and ``hat" attributes.]{
		\includegraphics[width=0.47\linewidth]{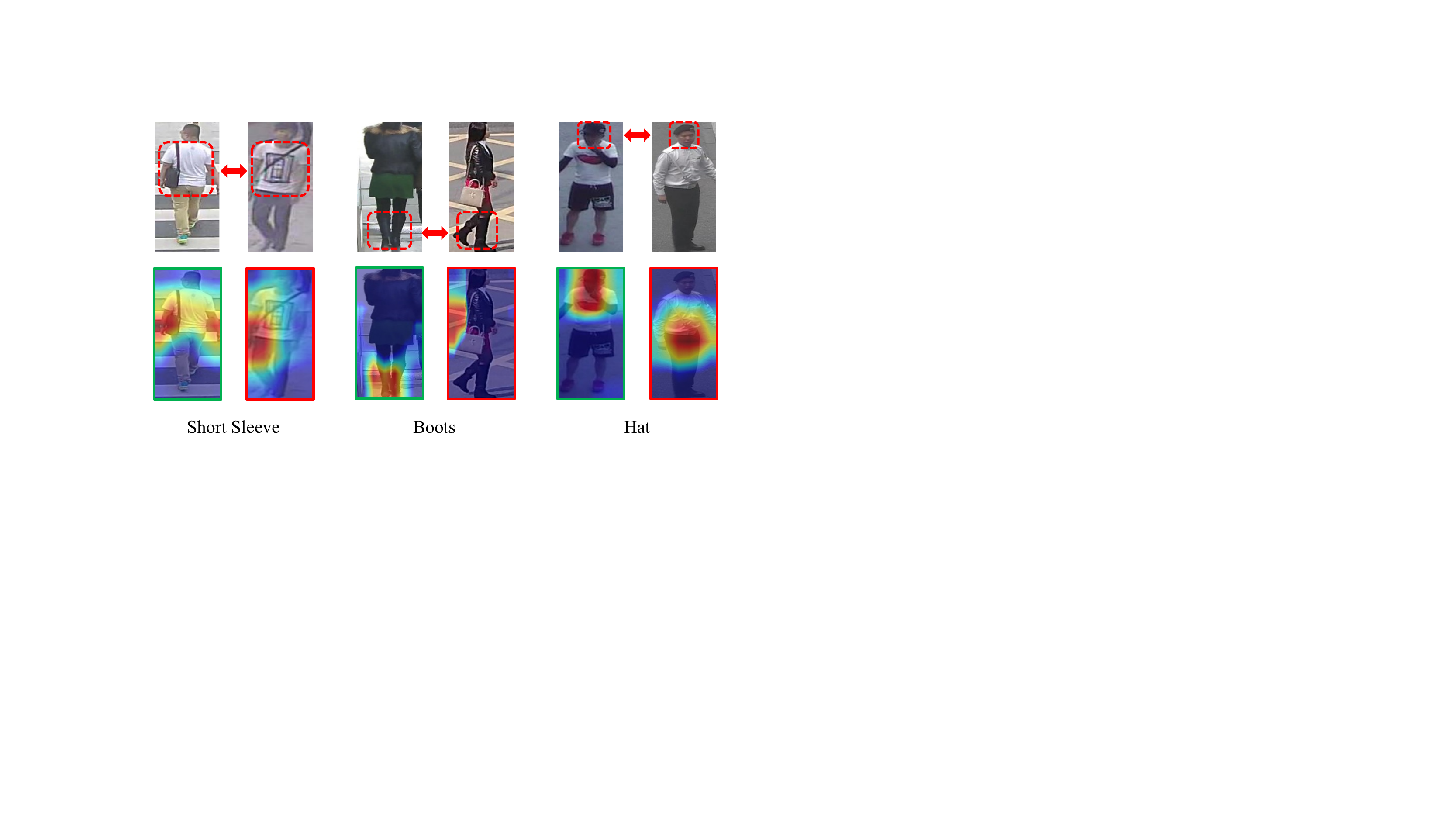}}
	\subfloat[Semantic consistency of different samples on the``hat" attribute.]{
		\includegraphics[width=0.53\linewidth]{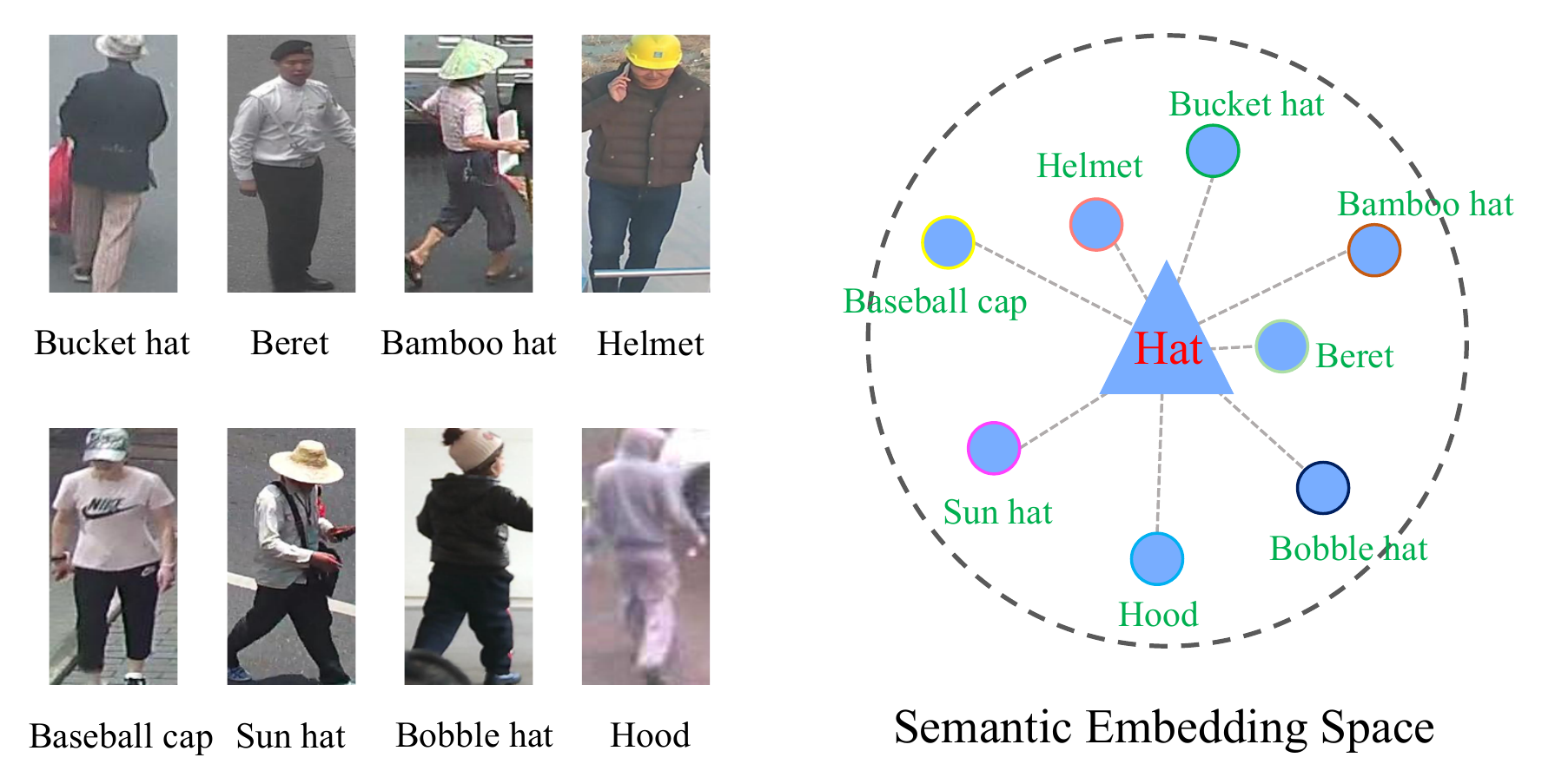}}
	\caption{{\bf Illustration of our main hypothesis on the spatial and semantic consistency}. In (a), CAMs of the baseline method in ``short sleeve", ``boots", and ``hat" attributes of the PA100K are visualized in the second row. Attribute-related regions of each attribute are plotted by the red dotted frame in the first row. Highlighted regions of the second CAM (with red boundary) of each attribute deviate from attribute-related regions severely, which are inconsistent with counterparts of the first CAM (with green boundary). In (b), we present several samples of the ``hat" attribute. Although these samples differ greatly in shape, size, and color, the intrinsic semantic features of the ``hat" attribute extracted by the model should remain unchanged. Best viewed in color.}
	\label{fig:problem}
	\vspace{-1em}
\end{figure*}

Pedestrian attribute recognition \cite{zhu2013pedestrian, wang2017attribute} aims to predict multiple human attributes, such as age, gender, and clothing, as semantic descriptions for a pedestrian image. Due to the ubiquitous application in surveillance scenarios \cite{wang2019parsurvey}, scene understanding \cite{shao2015deeply}, and human perception \cite{huang2020improve}, numerous methods \cite{deng2014pedestrian, li2015deepmar, li2016richly, wang2017attribute, li2018richly, liu2018localization, sarafianos2018deep, guo2019visual, tang2019Improving} have been proposed and significant progress has been made in the last decade.

Existing methods \cite{liu2017hydraplus, liu2018localization, sarafianos2018deep, tang2019Improving} mainly utilize the complicated network, such as Feature Pyramid Network (FPN), to enrich attribute representation from multi-level feature maps, and combine the attention mechanisms to precisely locate attribute-related regions. Recently, VAC \cite{guo2019visual} utilizes a human prior, that attention regions of random augmentations of the same image are consistent, to improve model robustness. The above methods \cite{liu2017hydraplus, liu2018localization, sarafianos2018deep, tang2019Improving} mainly emphasize learning discriminative attribute features from an individual image, instead of exploiting the relation between different pedestrian images of the same attribute. In contrast, our methods show that mining the inter-image relations between different images of the same attribute can significantly help the model locate attribute-related regions and extract inherent semantic features. We exploit inter-image relations from the perspective of spatial relation and semantic relation. 

For the inter-image spatial relation, we hypothesize that the spatial location of the same attribute is basically consistent between different pedestrian images, which is called SPAtial Consistency (SPAC) in this work. For example, the ``hat" attribute and the ``boots" attribute mostly appears at the top and bottom of the picture, respectively, which is shown in the first row of Figure \ref{fig:problem}(a). However, we observe that Class Activation Maps (CAMs) \cite{zhou2016learning} of the same attribute of the baseline method have significant location variations. Some examples are shown in the second row of Figure \ref{fig:problem}(a). 

These CAMs of the same attribute between different pedestrians are inconsistent, some of which (with red boundary) deviate seriously from attribute-related areas, no matter for the ``short sleeve", ``boots", or ``hat'' attributes. This phenomenon contradicts our spatial consistency hypothesis, and indicates that the baseline model easily inclines to focus on the background, irrelevant foreground, or a small part of attribute-related regions, which is called the ``spatial attention deviation problem" in this work.

For the inter-image semantic relation, inherent semantic features of the same attribute between different images should be consistent, which is called SEMantic Consistency (SEMC) in this work. For example, as illustrated in Figure \ref{fig:problem}(b), regardless of the difference in shape, size, and color between various samples, the intrinsic semantic features of the ``hat" attribute should remain basically unchanged. This property is also indispensable for learning discriminative features and obtaining a robust model.

To achieve the spatial and semantic consistency between pedestrian images of the same attribute, we propose a novel framework composed of the SPAC and SEMC module. Specifically, the SPAC module generates reliable spatial locations for each attribute and maintains a stable spatial memory to suppressing location shift, which is caused by overfitting or label noise. Based on precise spatial locations, the SEMC module extracts intrinsic semantic features and maintains a stable semantic memory to suppress the influence of irrelevant characteristics, such as shape, color, and size for the ``hat" attribute.

We make the following three contributions in this work:

\begin{itemize}
	\item We establish an effective consistency framework for pedestrian attribute recognition, which makes full use of inter-image spatial and semantic relations between images of the same attribute.
	\item We design spatial and semantic consistency modules to generate precise spatial attention regions and extract discriminative semantic features for each attribute.
	\item We confirm the efficacy of the proposed method by achieving state-of-the-art performance on three popular datasets including PA100K, PETA, and RAP.   
\end{itemize}

\section{Related Work} \label{related_work}

Pedestrian attribute recognition has witnessed a fast-growing development recently. Li \textit{et al.} \cite{li2015deepmar} first formulated pedestrian attribute recognition as a multi-label classification task and proposed the weighted sigmoid cross-entropy loss to alleviate the serious imbalance between positive samples and negative samples. To explore attribute context and correlation, the JRL network \cite{wang2017attribute} adopted Long-Shot-Term-Memory \cite{hochreiter1997long} to take the pedestrian attribute recognition task as a sequence prediction problem.

Attention mechanism \cite{liu2017hydraplus, liu2018localization, li2018pose, yang2019towards, TCLNet} has been widely used in pedestrian attribute recognition to locate attribute-related regions and learn discriminative feature representations. HydraPlus-Net \cite{liu2017hydraplus} with multi-directional attention modules was introduced to extract pixel-level features and semantic-level features, which were beneficial to locate fine-grained attributes. Based on CAM \cite{zhou2016learning} and EdgeBox \cite{zitnick2014edge}, Liu \textit{et al.} \cite{liu2018localization} proposed a Localization Guided Network to extract attribute-related local features. PGDM \cite{li2018pose} framework utilized a pre-trained human hose estimator and Spatial Transformer Networks (STNs) \cite{jaderberg2015spatial} to generate reliable attribute-related regions. 

\definecolor{green_me}{RGB}{153,204,153}
\definecolor{orange_me}{RGB}{255,204,153}

\begin{figure*}
	\centering
	\subfloat[Overview of the proposed framework.]{
		\includegraphics[width=0.45\linewidth ]{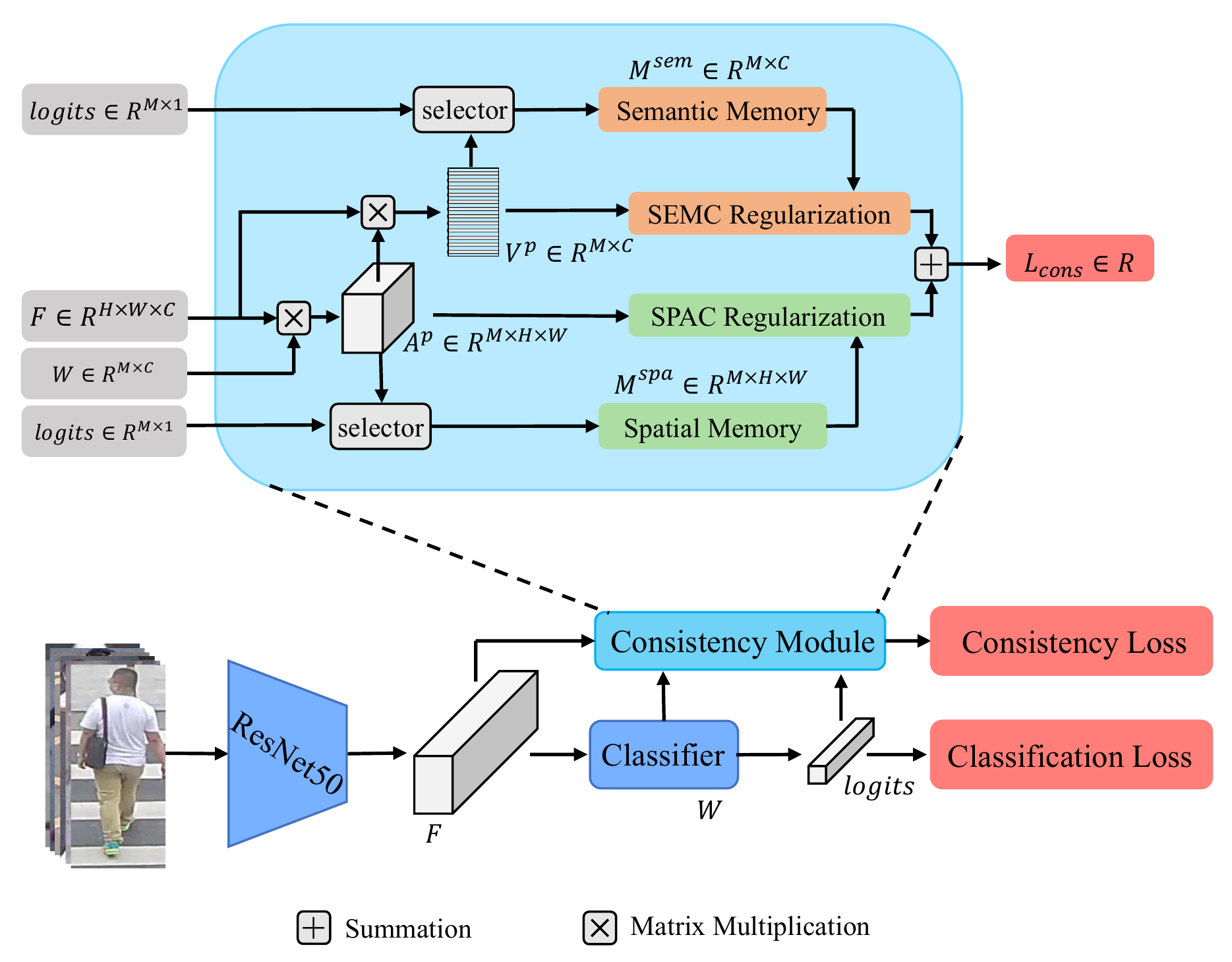}}
	\subfloat[Illustration of spatial and semantic consistency on ``hat" attribute.]{
		\includegraphics[width=0.54\linewidth ]{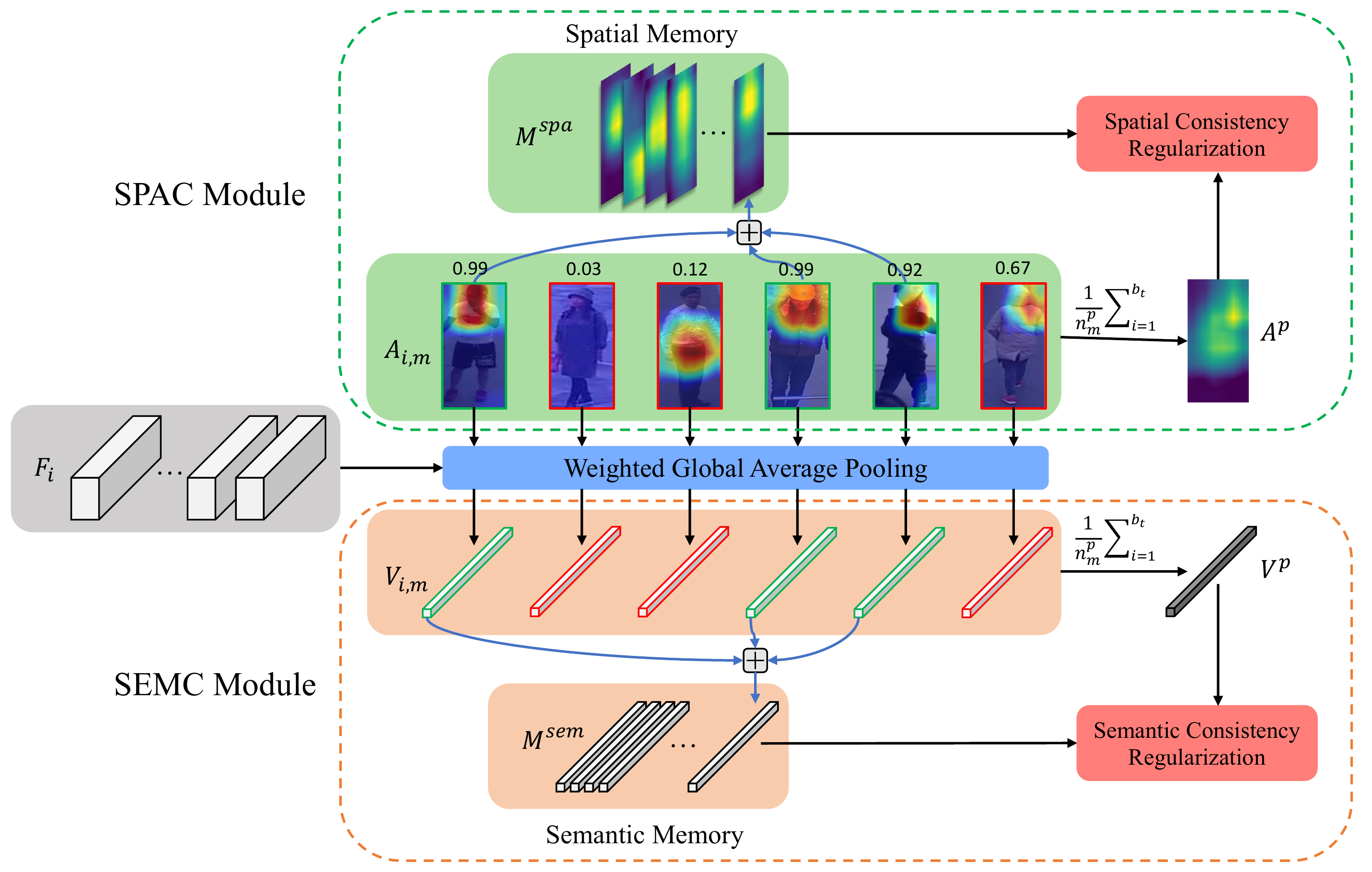}}
	\caption{{\bf Illustration of the proposed framework and consistency regularizations}. In (a), we describe two branch structures of the proposed framework and present the pipeline of the consistency module. In (b), we visually demonstrate how to construct spatial and semantic consistency regularizations from the SPAC module (\textcolor{green_me}{green shading \rule{0.25cm}{0.25cm}} ) and SEMC module (\textcolor{orange_me}{orange shading \rule{0.25cm}{0.25cm}} ) for the ``hat" attribute. For the SPAC module, only reliable CAMs of the ``hat" attribute are aggregated into the spatial memory $\bm{M}^{spa}$ as the supervision of spatial attention regions, but all CAMs $\bm{A}_{i, m}$ of the ``hat" attribute are utilized to compute the SPAC regularization. For the SEMC module, the semantic feature $\bm{V}_{i,m}$ is firstly extracted by the weighted global average pooling of the feature map $F^{i}$, and the weight parameters are the corresponding CAMs. After obtaining the semantic feature $\bm{V}_{i,m}$, SEMC memory $\bm{M}^{sem}$ and regularization are constructed as the same as that of the SPAC module. Prediction probabilities of the ``hat" attribute are listed above the CAMs. Best viewed in color.}
	\label{fig:model}
	\vspace{-1em}
\end{figure*}

Considering the discrimination of multi-scale feature maps and the effectiveness of deep supervisions,  WPAL \cite{yu2016weakly}, MsVAA \cite{sarafianos2018deep}, and ALM \cite{tang2019Improving} networks are proposed. Yu \textit{et al.} \cite{yu2016weakly} proposed the WPAL network, which introduced a weakly-supervised object detection technique into pedestrian attribute recognition. Sarafianos \textit{et al.} \cite{sarafianos2018deep} integrated attention mechanism into multi-scale feature maps and adopted a variant of focal loss to solve the imbalance between positive and negative samples of the attribute. ALM module \cite{tang2019Improving}, which was composed of a Squeeze-and-Excitation (SE) block \cite{hu2018squeeze} and a STN \cite{jaderberg2015spatial}, was applied to each layer of the Feature Pyramid Network (FPN) \cite{lin2017feature} to enhance attribute localization. Considering the visual attention regions were consistent between multiple augmentations of the same image, Guo \textit{et al.} \cite{guo2019visual} proposed an attention consistency loss to get robust attribute locations. In addition, a hierarchical feature embedding (HFE) \cite{yang2020hierarchical} framework was proposed to learn fine-grained feature embeddings by combining attribute and ID information. Different from previous methods, person ID information was utilized in the HFE framework, which was not provided on the pedestrian attribute recognition task.

Previous methods \cite{li2015deepmar, li2018pose, liu2017hydraplus, wang2017attribute, sarafianos2018deep, tang2019Improving, guo2019visual} mainly concentrated on generating precise attribute-related regions and learning to classify attributes from a single image individually. They neither considered the prior spatial structure knowledge of pedestrian attribute, nor exploited the inter-image relation between different pedestrian images of the same attribute. Whereas, both aspects are considered in our proposed method and introduced in Section \ref{spac} and \ref{semc}.

From the perspective of using inter-image information, the most related method is the JRL network \cite{wang2017attribute}. Based on global feature similarities, the JRL network utilized inter-image information by aggregating several similar pedestrian features to get the final prediction. Different from JRL, our method utilizes spatial and semantic local features of each attribute, and exploits the inter-image relation to construct consistency regularizations as supervision signals of the training process. From the perspective of consistency constraints, the most related method is the VAC model \cite{guo2019visual}, which aimed to make the global attention regions consistent between random augmentations of the same image. However, our method aligns the local attention regions between different pedestrian images of the same attribute. In addition, we also introduce the semantic consistency module to extract discriminative attribute features.

\section{Methods} \label{method}

In this section, we first introduce the baseline method. Then, we present the proposed consistency framework, which consists of a classification branch and a consistency branch. The classification branch is completely the same as the baseline network. The consistency branch is divided into spatial consistency module and semantic consistency module, which are introduced separately. The overview of the proposed framework is illustrated in Figure \ref{fig:model}(a), and intuitive elaborations of two consistency modules are shown in Figure \ref{fig:model}(b). Compared with the baseline method, the proposed method does not introduce extra learnable parameters.

\subsection{Baseline Method}
Given a dataset $\mathbb{D}=\{(\bm{X}_i, \bm{y}_i)\ | \ i=1, 2, \ldots, N \}$, pedestrian attribute recognition aims to predict multiple attribute $\bm{y}_i \in \{0, 1\}^{M}$ to $i$-th pedestrian image, where $N$, $M$ denotes the number of images and attributes respectively. The zeros and ones in the attribute vector $\bm{y}_i$ indicate the absence and presence of the corresponding attributes in the pedestrian image.

Following \cite{li2018richly, tang2019Improving, sarafianos2018deep, guo2019visual}, we formulate pedestrian attribute recognition as a multi-label classification task, and multiple binary classifiers with sigmoid functions \cite{li2015deepmar, li2018richly} are adopted. Binary cross-entropy loss is used as the optimization target:
\begin{align}
	L_{cls} & = \frac{1}{N} \sum_{i=1}^{N} \sum_{j=1}^{M} y_{i,j} \log(p_{i,j})  + (1 - y_{i,j})\log(1 - p_{i,j}) ,
	\label{eq:celoss}
\end{align}
where $p_{i,j} = \sigma(z_{i,j})$ is the prediction probability of the classifier output logits $z_{i,j}$, and $\sigma(x) = 1/(1+ e^{-x})$ is the sigmoid function.

\subsection{Spatial Consistency Module} \label{spac}

In this section, we propose the SPAtial Consistency (SPAC) module combined with spatial consistency regularization to tackle the spatial attention region deviation problem.  

SPAC module takes feature map $\bm{F}_{i} \in \mathbb{R}^{H \times W \times C}$, classifier weight $\bm{W} \in \mathbb{R}^{M \times C}$, and logits $\bm{z}_{i} \in \mathbb{R}^{M \times 1}$ as inputs, where $\bm{F}_{i}$ is the output of the backbone network (ResNet-50 \cite{he2016deep} used in our work) for image $\bm{X}_i$, and $H, W, C$ represent height, width, and channel dimension of feature map respectively. Inspired by the Class Activation Map (CAM) \cite{zhou2016learning}, we first obtain spatial attention maps $\bm{A}_{i, m} \in \mathbb{R}^{H \times W} $ of $m$-th attribute for image $\bm{X}_{i}$ as follows:
\begin{align}
	\bm{A}_{i,m}(x, y) & = \sum_{c=1}^{C} \bm{W}_{m, c} \bm{F}_{i, c}(x, y), & m \in \{ 1, 2, \ldots , M \}   \label{eq:attn_map},
\end{align}
where $\bm{W}_{m, c}$ denotes the $c$-th element of $m$-th classifier weight, and $\bm{F}_{i, c}(x, y)$ indicates the spatial location $(x, y)$ of the $c$-th channel  in the feature map $\bm{F}_{i}$. After getting the spatial attention regions of each attribute for every image in a random batch $b_{t}$, we adopt the selector --- an indicator function takes logits $z_{i, m}$ and ground truth label $y_{i, m}$ as inputs --- to aggregate the attention maps of qualified positive samples \footnote{We use ``positive samples" to represent images that contain target attribute, and ``negative samples" to represent images that do not contain target attribute.} of the $m$-th attribute by:
\begin{align}
	\bm{A}^{q}_{m}(x, y) & = \frac{1}{n_{m}^{q}} \sum_{i=1}^{b_{t}} \mathds{1}_{\{\sigma(z_{i, m}) >\tau, \ y_{i, m} = 1 \}} \bm{A}_{i, m}(x, y) ,\label{eq:tau_a}
\end{align}
where $\bm{A}^{q} = \{ {\bm{A}^{q}_m} | m \in {1, 2, \ldots, M} \} \in \mathbb{R}^{M \times H \times W}$ denotes the attention map aggregation of qualified positive samples for each attribute, and $n_m^{q} = \sum_{i=1}^{b_{t}} \mathds{1}_{\{\sigma(z_{i, m}) >\tau, \ y_{i, m} = 1 \}}$ indicates the number of qualified positive samples of $m$-th attribute in a random batch $b_{t}$. Prediction probabilities of these qualified positive samples are required to be higher than a confidence threshold $\tau$ (default as 0.9). The robustness of hyper-parameter $\tau$ is validated by the experiments in Figure \ref{tab:hyper_tau} . 

Through strict selection, $\bm{A}^{q}_{m}$ can be regarded as reliable spatial locations for $m$-th attribute on current batch. To save the reliable spatial location in every batch, the spatial attention maps $\bm{A}^{q}_{m}$ of each attribute are normalized and aggregated into spatial memory $\bm{M}^{spa}=\{\bm{M}^{spa}_{m}| m \in {1, 2, \ldots, M}\} \in \mathbb{R}^{M \times H \times W}$ in a momentum updated way to decrease spatial location variation, \ie,
\begin{align}
	\bm{M}^{spa}_{m} & \leftarrow ( 1 - \alpha ) \times \bm{\bar{M}}^{spa}_{m} + \alpha \times \bm{\bar{A}}^{q}_{m}, \label{eq:alpha_a}
\end{align}
where $\bm{\bar{M}}^{spa}_{m} = \bm{M}^{spa}_{m} /\ \| \bm{M}^{spa}_{m} \|_{2}$,  $ \bm{\bar{A}}^{q}_{m} = \bm{A}^{q}_{m} /\ \| \bm{A}^{q}_{m} \|_{2} $, and $\alpha \in (0,1]$ is a momentum coefficient. The effect of momentum coefficient $\alpha$ is demonstrated in Figure \ref{tab:hyper_alpha}.


As shown in Figure \ref{fig:model}(b), due to overfitting and label noise, spatial attention regions of the ``hat" attribute deviate from attribute-related regions severely. Model inclines to focus on the background, irrelevant foreground, and a small part of the attribute-related areas. Thus, spatial memory $\bm{M}^{spa}$, which retains reliable and stable spatial location regions of each attribute, can be taken as the supervision of attribute-related regions to correct the spatial attention deviation. Therefore, based on SPAC module, we propose a spatial consistency regularization $L_{spac}$ by calculating the $l_{1}$-distance between the spatial memory $\bm{M}^{spa}_{m}$ and the spatial attention map $\bm{A}^{p}_{m}$ :
\begin{align}
	L_{spac} & =  \frac{1}{M} \sum_{m=1}^{M} \| \bm{\bar{A}}^{p}_{m} - \bm{\bar{M}}^{spa}_{m}\|_{1}, \\
	\bm{A}^{p}_{m}(x, y) & = \frac{1}{n_{m}^{p}} \sum_{i=1}^{b_{t}} \mathds{1}_{\{y_{i, m} = 1 \}} \bm{A}_{i, m}(x, y), 
\end{align}
where $\bm{\bar{A}}^{p}_{m} = \bm{A}^{p}_{m} /\ \|\bm{A}^{p}_{m}\|_{2}$, $n_m^{p} = \sum_{i=1}^{bt} \mathds{1}_{\{\ y_{i, m} = 1 \}}$, and $b_{t}$ indicates the batch size. To take all positive samples into consideration, $\bm{A}^{p}_{m}$ is formulated by averaging spatial attention regions of all positive samples of the $m$-th attribute in a random batch. Please note the difference between indicator functions of $\bm{A}^{q}_{m}$ and $\bm{A}^{p}_{m}$.

Overall, to fully utilize the inter-image spatial relation and address the spatial attention region deviation problem, the SPAC module is proposed to extract reliable attribute attention regions $\bm{A}^{q}_{m}$ to update spatial memory and adopt the $l_1$-distance to align spatial attention regions $\bm{A}^{p}_{m}$ with spatial memory $\bm{M}^{spa}_{m}$. Considering the soft weights used in $\bm{A}^{p}_{m}$ and $\bm{M}^{spa}_{m}$, we name this method as $SSC_{soft}$.

\subsection{Semantic Consistency Module} \label{semc}

Although the SPAC module considers the inter-image spatial relation that attention regions of different images of the same attribute are consistent, inter-image semantic relation has not been utilized, \ie, intrinsic semantic features of the same attribute are consistent between different images. For example, whether the sample is a beret, helmet, bucket hat, or baseball cap, intrinsic semantic features of the ``hat" attribute should be consistent. Thus, based on the SPAC module, we propose SEMantic Consistency (SEMC) module to extract intrinsic and discriminative semantic features for each attribute.

According to Equation \ref{eq:attn_map}, we first compute the spatial attention map $\bm{A}_{i, m}$ of $m$-th attribute for image $\bm{X}_i$ to obtain attribute-related regions. Then semantic feature vector $\bm{V}_{i, m} \in \mathbb{R}^{C \times 1}$ can be constructed by weighted global average pooling as:
\begin{align}
	\bm{V}_{i, m} & = \frac{1}{H \times W} \sum_{x=1}^{H} \sum_{y=1}^{W} A_{i, m}(x, y)F_{i}(x, y) ,
\end{align}
where $F_{i}(x, y) \in \mathbb{R}^{C \times 1}$ is a spatial feature vector of position $(x, y)$. To provide a consistent supervision for semantic features, we maintain a stable and discriminative semantic memory $\bm{M}^{sem} = \{\bm{M}^{sem}_{m} | m \in {1, 2, \ldots, M} \} \in \mathbb{R}^{M \times C}$ for each attribute. 
The selector is adopted in the same way as the SPAC module to aggregate reliable semantic features $V^{q}_{m} \in \mathbb{R}^{C \times 1}$ into $\bm{M}^{sem}_{m}$ in a momentum updated way: 
\begin{align}
	\bm{V}^{q}_{m} = & \frac{1}{n_{m}^{q}} \sum_{i=1}^{b_{t}} \mathds{1}_{\{\sigma(z_{i, m}) >\tau, \ y_{i, m} = 1 \}} \bm{V}_{i, m} \label{eq:tau_v},\\
	\bm{M}^{sem}_{m} & \leftarrow ( 1 - \alpha ) \times \bm{\bar{M}}^{sem}_{m} + \alpha \times \bm{\bar{V}}^{q}_{m} \label{eq:alpha_v},
\end{align}
where $\bm{\bar{V}}^{q}_{m} = \bm{V}^{q}_{m} /\ \|\bm{\bar{V}}^{q}_{m}\|_{2}$ , $n_{m}^{q} = \sum_{i=1}^{b_{t}} \mathds{1}_{\{\sigma(z_{i, m}) >\tau,\ y_{i, m} = 1 \}}$, and $\alpha$ is momentum coefficient as same as that of the SPAC module in Equation \ref{eq:alpha_a} .

Finally, we design a semantic consistency regularization by computing the $l_{1}$-distance between the semantic memory $\bm{M}^{sem}_{m}$ and attribute semantic feature $\bm{V}^{p}_{m}$ of all positive samples, which is defined as: 
\begin{align}
	L_{semc} & =  \frac{1}{M} \sum_{m=1}^{M} \| \bm{\bar{V}}^{p}_{m} - \bm{\bar{M}}^{sem}_{m}\|_{1}, \\
	\bm{V}^{p}_{m} & = \frac{1}{n_{m}^{p}} \sum_{i=1}^{b_{t}} \mathds{1}_{\{ y_{i, m} = 1 \}} \bm{V}_{i,m},
\end{align}
where $\bm{\bar{V}}^{p}_{m} = \bm{V}^{p}_{m} /\ \|\bm{V}^{p}_{m}\|_{2} $, $\bm{\bar{M}}^{sem}_{m} = \bm{M}^{sem}_{m} /\ \|\bm{M}^{sem}_{m}\|_{2}$, $n_{m}^{q} = \sum_{i=1}^{b_{t}} \mathds{1}_{\{y_{i, m} = 1 \}}$, and the semantic consistency regularization is imposed on semantic features of all positive samples of the $m$-th attribute. 

By bridging the gap between semantic features of different samples of the same attribute, the SEMC module can extract intrinsic and discriminative semantic features for each attribute and eliminate the interference of attribute-irrelevant characteristics (such as shape, size, and color in the ``hat" attribute).

\subsection{Loss Function}

As commonly adopted in most existing methods \cite{sarafianos2018deep, guo2019visual, tang2019Improving}, the weighted binary cross-entropy loss is also utilized in the  classification branch of the proposed method as classification loss, which is formulated as :
\begin{gather}
	L_{cls} = \frac{1}{N} \sum_{i=1}^{N} \sum_{j=1}^{M} \omega_{i,j}(y_{i,j} \log(p_{i,j})  + (1 - y_{i,j})\log(1 - p_{i,j})), \\
	\omega_{i,j} =  y_{i,j}e^{1 - r_{j}}  + (1 - y_{i,j})e^{r_{j}},
	\label{eq:weight_celoss}
\end{gather}
where $r_{j} $ is the positive sample ratio of $j$-$th$ attribute in the training set.

The final loss function $L$ is a weighted summation of the classification loss, SPAC regularization, and SEMC regularization:
\begin{align}
	L & = L_{cls} +  \mathds{1}_{\{ e > i_{e} \}} (\lambda_{1} L_{spac} + \lambda_{2} L_{semc}), \label{eq:final_loss}
\end{align}
where $\lambda_{1} = 1, \ \lambda_{2} = 0.1$ is set as default in all experiments if not specially specified. Current epoch number in the training stage is indicated by $e \in \{ 0, \cdots 30 \} $, and initial epoch $i_{e}$ is used to ensure reliable consistency memory and effective consistency regularization.

\section{Experiments} \label{exps}

\renewcommand{\thefootnote}{\fnsymbol{footnote}}
\begin{table*}[h]
\centering
\caption{{\bf Performance comparison with state-of-the-art methods on the PETA, RAP, and PA100K datasets.} Five metrics, mean accuracy (mA), accuracy (Accu), precision (Prec), recall (Recall), F1 are evaluated. To make a fair comparison, we also report our reimplementation performance for the MsVAA, VAC, and ALM methods. The \textcolor{red}{first} and \textcolor{blue}{second} highest scores are represented by red font and blue font respectively. The difference between $SSC_{soft}$, $SSC_{hard}$, and $SSC_{fix}$ lies in the implementation of $M^{spa}$ and $A^{q}$, and we detail it in Section \ref{ablation} .}
\resizebox{\linewidth}{!}{
\begin{tabular}{l|l|ccccc|ccccc|ccccc}
\toprule
\multirow{2}{*}{Method} & \multirow{2}{*}{Backbone} & \multicolumn{5}{c|}{PETA} & \multicolumn{5}{c|}{PA100K} &\multicolumn{5}{c}{RAP} \\ \cline{3-17} 
	&	& mA & Accu & Prec & Recall & F1 & mA & Accu & Prec & Recall & F1 & mA & Accu & Prec & Recall & F1 \\ \midrule \midrule
DeepMAR \cite{li2015deepmar} & CaffeNet & 82.89 & 75.07 & 83.68 & 83.14 & 83.41 & 72.70 & 70.39 & 82.24 & 80.42 & 81.32 & 73.79 & 62.02 & 74.92 & 76.21 & 75.56 \\
HPNet\cite{liu2017hydraplus} & InceptionNet & 81.77 & 76.13 & 84.92 & 83.24 & 84.07 & 74.21 & 72.19 & 82.97 & 82.09 & 82.53 & 76.12 & 65.39 & 77.33 & 78.79 & 78.05 \\
JRL \cite{wang2017attribute} & AlexNet & 85.67 & -- & 86.03 & 85.34 & 85.42 &--&--&--&--&--& 77.81 & -- & 78.11 & 78.98 & 78.58 \\
LGNet \cite{liu2018localization} & Inception-V2 &--&--&--&--&--& 76.96 & 75.55 & 86.99 & 83.17 & 85.04 & 78.68 & 68.00 & 80.36 & 79.82 & 80.09 \\
PGDM \cite{li2018pose} & CaffeNet & 82.97 & 78.08 & 86.86 & 84.68 & 85.76 & 74.95 & 73.08 & 84.36 & 82.24 & 83.29 & 74.31 & 64.57 & 78.86 & 75.90 & 77.35\\ \midrule
MsVAA\cite{sarafianos2018deep} & ResNet101 & 84.59 & 78.56 & 86.79 & 86.12 & 86.46 &--&--&--&--&--&--&--&--&--&--\\

VAC \cite{guo2019visual} & ResNet50 &--&--&--&--&--& 79.16 & \textcolor{red}{\bf{79.44}} & \textcolor{blue}{\bf{88.97}} & 86.26 & \textcolor{red}{\bf{87.59}} &--&--&--&--&-- \\
ALM \cite{tang2019Improving}  & BN-Inception & \textcolor{blue}{\bf{86.30}} & \textcolor{red}{\bf{79.52}} & 85.65 & 88.09 & \textcolor{blue}{\bf{86.85}} & 80.68 & 77.08 & 84.21 & 88.84 & 86.46 & 81.87 & 68.17 & 74.71 & 86.48 & 80.16 \\ 
\midrule
MsVAA\cite{sarafianos2018deep} \footnotemark[1] & ResNet50 & 84.35 & 78.69 & 87.27 & 85.51 & 86.09 & 80.10 & 76.98 & 86.26 & 85.62 & 85.50 & 79.75 & 65.74 & 77.69 & 78.99 & 77.93 \\
VAC \cite{guo2019visual} \footnotemark[1] & ResNet50 & 83.63 & 78.94 & \textcolor{blue}{\bf{87.63}} & 85.45 & 86.23 & 79.04 & 78.95 & 88.41 & 86.07 & 86.83 & 78.47 & 68.55 & \textcolor{blue}{\bf{81.05}} & 79.79 & 80.02 \\  
ALM \cite{tang2019Improving} \footnotemark[1] & ResNet50 & 85.50 & 78.37 & 83.76 & \textcolor{blue}{\bf{89.13}} & 86.04 & 79.26 & 78.64 & 87.33 & 86.73 & 86.64 & 81.16 & 67.35 & 74.97 & 85.36 & 79.39\\ 
\midrule
Baseline & ResNet50 & 81.15 & 77.96 & \textcolor{red}{\bf{88.19}} & 83.77 & 85.56 & 78.53 & 78.87 & \textcolor{red}{\bf{88.99}} & 85.38 & 86.34 & 76.09 & \textcolor{red}{\bf{68.66}} & \textcolor{red}{\bf{83.74}} & 77.44 & 79.50 \\
$SSC_{soft}$ & ResNet50 & \textcolor{red}{\bf{86.52}} & 78.95 & 86.02 & 87.12 & \textcolor{red}{\bf{86.99}} & \textcolor{red}{\bf{81.87}} & \textcolor{blue}{\bf{78.89}} & 85.98 & \textcolor{red}{\bf{89.10}} & 86.87 & \textcolor{blue}{\bf{82.77}} & \textcolor{blue}{\bf{68.37}} & 75.05 & \textcolor{blue}{\bf{87.49}} & \textcolor{red}{\bf{80.43}}  \\
$SSC_{hard}$ & ResNet50 & 85.92 & 78.53 & 86.31 & 86.23 & 85.96 & 81.02 & 78.42 & 86.39 & 87.55 & 86.55 & 82.14 & 68.16 & 77.87 & 82.88 & 79.87 \\
$SSC_{fix}$ & ResNet50 & 86.07 & \textcolor{blue}{\bf{79.23}} & 84.58 & \textcolor{red}{\bf{89.26}} & 86.54 & \textcolor{blue}{\bf{81.70}} & 78.85 & 85.80 & \textcolor{blue}{\bf{88.92}} & \textcolor{blue}{\bf{86.89}} & \textcolor{red}{\bf{82.83}} & 68.16 & 74.74 & \textcolor{red}{\bf{87.54}} & \textcolor{blue}{\bf{80.27}}\\\bottomrule
\end{tabular}}
\vspace{-1em}
\label{tab:peta_rap_perf}
\end{table*}

\subsection{Datasets and Evaluation Metrics} 

\noindent {\bf Datasets.} We perform experiments on the PETA \cite{deng2014pedestrian}, RAP \cite{li2016richly}, and PA100K \cite{liu2017hydraplus}. The PEdesTrian Attribute (PETA) dataset \cite{deng2014pedestrian} is collected from 10 small-scale person datasets and consists of 19,000 person images, which is divided into 9500 images for the training set, 1900 for the validation set, and 7600 for the test set. Each image is labeled with 61 binary attributes and 4 multi-class attributes. We follow the common experimental protocol \cite{li2018richly, sarafianos2018deep, tang2019Improving}, and only 35 attributes whose positive ratios are higher than 5\% are used for evaluation. The Richly Annotated Pedestrian (RAP) attribute dataset \cite{li2016richly} consists of 33,268 images for training and 8,317 images for testing, a total of 41,585 images extracted from 26 indoor surveillance cameras. Each image is labeled with 69 binary attributes and 3 multi-class attributes. Following the official protocol \cite{li2016richly}, 51 binary attributes are adopted to evaluate the recognition performance. The PA100K dataset \cite{liu2017hydraplus} consists of 100,000 pedestrian images and is split into training, validation, and test sets with a ratio of 8:1:1. Each image is described with 26 commonly used attributes. Considering the identical pedestrian identities between training set and test set \cite{jia2020rethinking} on the RAP and PETA, performance on the largest dataset PA100K is more convincible.

\noindent {\bf Evaluation Protocal.}
Two types of metrics, \ie, a label-based metric and four instance-based metrics, are adopted to evaluate attribute recognition performance \cite{li2018richly}. For the label-based metric, we compute the mean value of classification accuracy of positive samples and negative samples as the metric for each attribute. Then we take an average over all attributes as mean accuracy. For instance-based metrics, accuracy, precision, recall, and F1-score are used.

\subsection{Implementation Details} \label{details}
The proposed method is implemented with PyTorch and trained in an end-to-end manner. We adopt ResNet50 \cite{he2016deep} as the backbone network to extract pedestrian image features for a fair comparison. Pedestrian images are resized to $256 \times 192$ as inputs. Random horizontal mirroring, padding, and random crop are used as augmentations. Adam is employed for training with the weight decay of 0.0005. The initial learning rate equals 0.0001, and the batch size is set to 64. Plateau learning rate scheduler is used with reduction factor 0.1 and loss patience 4. The total epoch number of the training stage is 30. Momentum coefficient $\alpha = 0.9$, confidence threshold $\tau = 0.9$ by default. To obtain the stable and reliable spatial memory $\bm{M}^{spa}_{m}$ and semantic memory $\bm{M}^{sem}_{m}$, consistency regularizations are added to the classification loss after epoch 4, \ie, $i_{e} = 4$ in Equation \ref{eq:final_loss} .

\subsection{Comparison to the State of the Arts}

\footnotetext[1]{Results are reimplemented in the same setting for a fair comparison.}

\begin{table*}[h]
\centering
\caption{{\bf Ablation study of each component of our method on the PETA, PA100K, RAP}. Performance improved by spatial consistency (SPAC) and semantic consistency (SEMC) regularizations validates the effectiveness of our methods. We use $SSC_{soft}$ model as default. }
\resizebox{\linewidth}{!}{
\begin{tabular}{c|c|c|ccccc|ccccc|ccccc}
\toprule
\multicolumn{3}{c|}{Method} & \multicolumn{5}{c|}{PETA} & \multicolumn{5}{c|}{PA100K} & \multicolumn{5}{c}{RAP}\\ \cline{1-18}  
	 SEMC & SPAC & Weighted Loss & mA & Accu & Prec & Recall & F1 & mA & Accu & Prec & Recall & F1 & mA & Accu & Prec & Recall & F1\\ \midrule \midrule
- & - & - & 81.15 & 77.96 & 88.19 & 83.77 & 85.56 & 78.53 & 78.87 & \bf{88.99} & 85.38 & 86.75 & 76.09 & 68.66 & \bf{83.74} & 77.44 & 80.06 \\
$\checkmark$ & - &  - & 82.34 & 78.51 & \bf{88.29} & 84.36 & 85.94 & 78.63 & 78.68 & 88.63 & 85.52 & 86.64 & 76.45 & 68.58 & 83.08 & 77.81 & 79.97 \\
- & $\checkmark$ & - & 84.08 & 78.85 & 87.66 & 85.32 & 86.19 & 79.15 & 78.62 & 88.24 & 85.71 & 86.57 &78.55 & 68.60 & 82.09 & 79.34 & 79.99 \\
- & - & $\checkmark$ & 84.17 & 78.81 & 87.30 & 85.58 & 86.15 & 79.59 & 78.86 & 87.70 & 86.65 & 86.77 & 79.46 & 66.55 & 78.39 & 79.62 & 78.58 \\
$\checkmark$ & $\checkmark$ & -  & 84.90 & 78.49 & 86.44 & 85.90 & 85.91 & 80.09 & \bf{79.11} & 88.37 & 86.33 & \bf{86.95} & 80.26 & \bf{68.77} & 80.64 & 80.07 & 80.29 \\
$\checkmark$ & $\checkmark$ & $\checkmark$ & \bf{86.52} & \bf{78.95} & 86.02 & \bf{87.12} & \bf{86.99} & \bf{81.87} & 78.86 & 85.98 & \bf{89.10} & 86.87 & \bf{82.77} & 68.37 & 75.05 & \bf{87.49} & \bf{80.43} \\\bottomrule
\end{tabular}}
\vspace{-1em}
\label{tab:ablation}
\end{table*}

In Table \ref{tab:peta_rap_perf}, we compare the performance of the proposed methods with several existing algorithms on the PETA, RAP, and PA100K. For a fair comparison, besides the performance reported by the papers \cite{sarafianos2018deep, guo2019visual, tang2019Improving}, we also report the performance of our reimplements based on the same setting described in Section \ref{details}. 

Compared with the performance reported by the paper of MsVAA \cite{sarafianos2018deep}, VAC \cite{guo2019visual}, and ALM \cite{tang2019Improving} methods, the $SSC_{soft}$ model achieves better performance on the PETA, PA100K, and RAP without increasing learnable parameters. Compared to the MsVAA model adopted ResNet101 as the backbone network, we achieve 1.93\% and 0.53\% performance improvements in mA and F1 on the PETA dataset. Compared to the ALM model, which utilizes the complicated combination of FPN, STN and SE modules introducing extra 17\% parameters, the $SSC_{soft}$ method achieves 0.22\%, 1.19\%, and 0.9\% performance improvements in mA on three popular datasets. Besides, compared with the performance achieved by our reimplementation of the MsVAA, VAC, and ALM methods, the performance of the $SSC_{soft}$ method has significant improvements from 1.02\% to 4.30\% in mA on the PETA, PA100K, and RAP, which fully demonstrates the effectiveness of our method.

It can be noticed that the proposed spatial and semantic consistency method substantially outperforms the visual attention consistency (VAC) method \cite{guo2019visual}. The VAC method hypothesis that global attention regions of random augmentations of the same image are consistent. However, the VAC method focuses on global attention regions of an individual image and cannot generate precise local attention regions for each fine-grained attribute. In addition, for a pair of augmentations of the same image, if global attention regions of one augmentation are precise, the VAC method can improve the performance by aligning the global attention regions of another augmentation with these of the current one. However, if attention regions of both augmentations are inaccurate, the VAC method cannot solve the attention region deviation problem, which can be addressed by our proposed method.

\subsection{Ablation Study and Discussion} \label{ablation}

In this section, we first investigate the effect of the SPAC and SEMC module by conducting analytical experiments on all three datasets. We then introduce two variants of our methods to demonstrate the effectiveness of spatial and semantic consistency regularizations. Quantitative performance improvements of each attribute on three datasets are presented in the supplementary material.

As shown in Table \ref{tab:ablation}, compared to the baseline method, we have the following observations. First, adopting the SEMC module alone can hardly bring performance improvement. The results prove that, without correct attention regions, attribute semantic features lack discrimination and contain more noise, which is in line with the intuitive hypothesis. Second, adopting the SPAC module can directly bring 2.93\%, 0.62\%, 2.46\% performance improvements in mA on the PETA, PA100K, and RAP, respectively. This improved performance demonstrates that spatial consistency regularization is beneficial for locating the attribute-related regions.  Third, when the SPAC module and SEMC module are jointly adopted, our method improves the performance over the baseline model by 3.75\%, 1.56\%, 4.17\% in mA on the PETA, PA100K, and RAP.

To further validate the reasonableness of proposed spatial and semantic consistency regularizations, we implement our method with two variants $SSC_{hard}$ and $SSC_{fix}$. For the $SSC_{hard}$ method, we change the $\bm{A}^{q}$ in Equation \ref{eq:tau_a} and $\bm{M}^{spa}$ in Equation \ref{eq:alpha_a} of SPAC module from soft attention maps to binary (hard) attention maps based on a threshold $th_{hard}=0$. For the $SSC_{fix}$ method, we first train a baseline model and obtain the qualified CAMs $\bm{A}^{q}$ of positive samples for each attribute according to Equation \ref{eq:tau_a}. Then, we fix $\bm{M}^{spa}$ as $\bm{A}^{q}$ instead of momentum updating to training a new model $SSC_{fix}$. The experimental results of two variants are listed in Table \ref{tab:peta_rap_perf}. Although method $SSC_{hard}$ assigns the same weight to each pixel of the region of interest, which is not as flexible as $SSC_{soft}$ and achieves slightly reduced performance, it still achieves competitive performance on PA100k and RAP. Since the $SSC_{soft}$ and $SSC_{fix}$ method can get reliable and accurate spatial attention regions $\bm{M}^{spa}$, they both achieve the state-of-the-art performance. However, compared to the $SSC_{fix}$ method, $SSC_{soft}$ with the momentum updated memory can avoid a two-stage training process and is more suitable for industry application.

\subsection{Effects of SPAC and SEMC Module}

Spatial and semantic consistency regularizations are two complementary and indispensable parts of a powerful model. The SPAC module can enhance the localization capability of the backbone network without being disturbed by overfitting and label noise. Based on the precise spatial attention regions of attributes, the backbone network further benefits from the SEMC module to extract intrinsic and discriminative semantic features.

\begin{figure}
	\centering
	\subfloat[Visualization of spatial attention regions $\bm{A}_{i, m}$ between the baseline method (red boundary) and the proposed method (green boundary).]{
		\includegraphics[width=1\linewidth ]{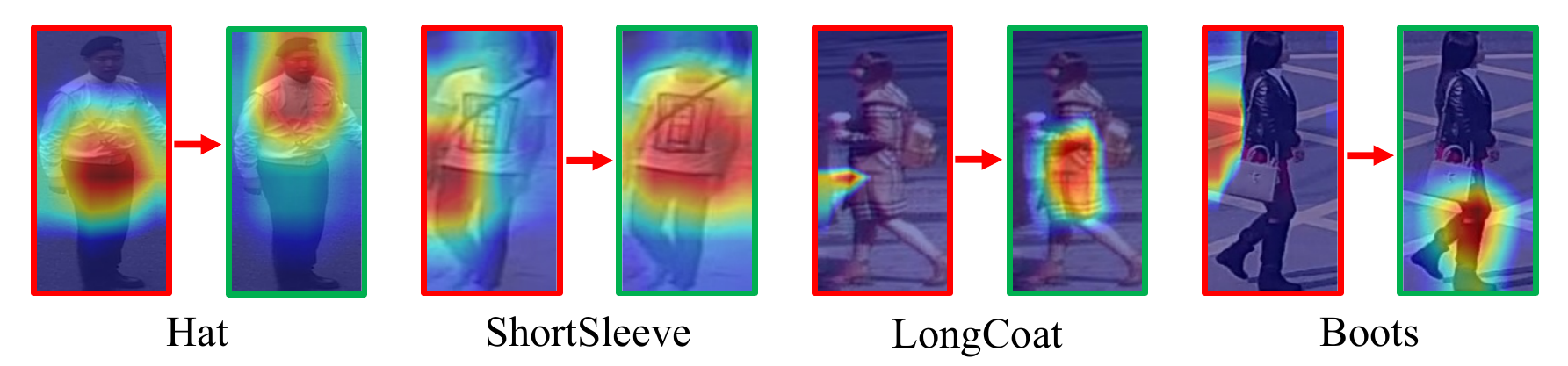}}
		
	\subfloat[Comparison of the similarity distribution of spatial attention regions $\bm{A}_{i, m}$.]{
		\includegraphics[width=1\linewidth ]{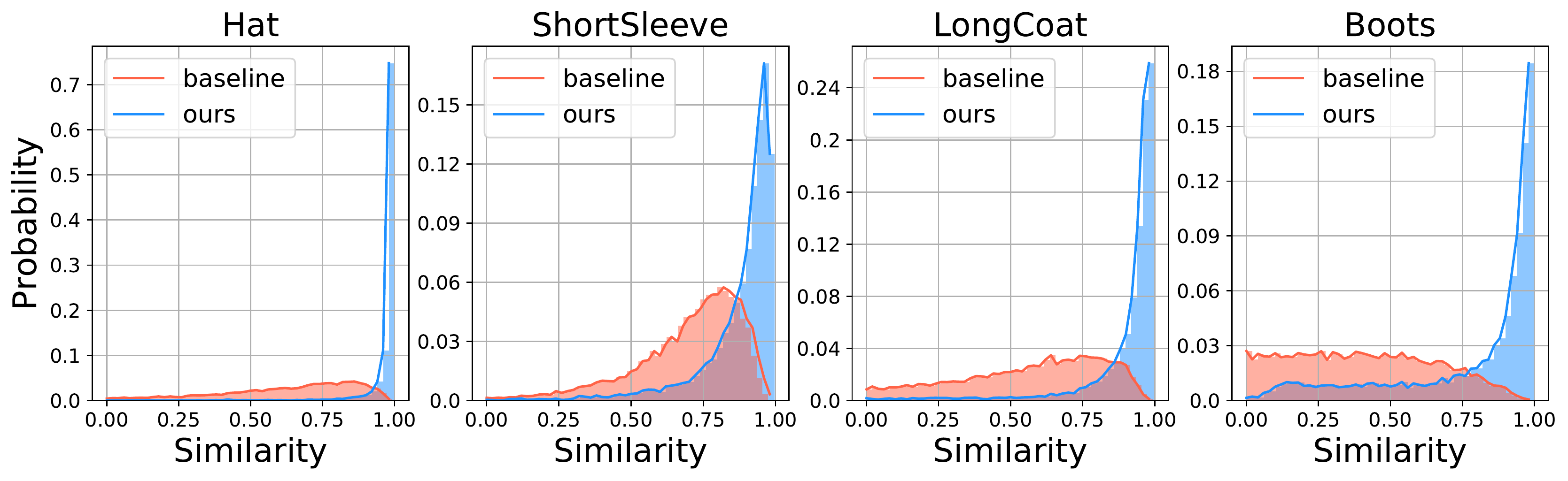}}
		
	\subfloat[Comparison of the similarity distribution of semantic features $\bm{V}_{i, m}$.]{
		\includegraphics[width=1\linewidth ]{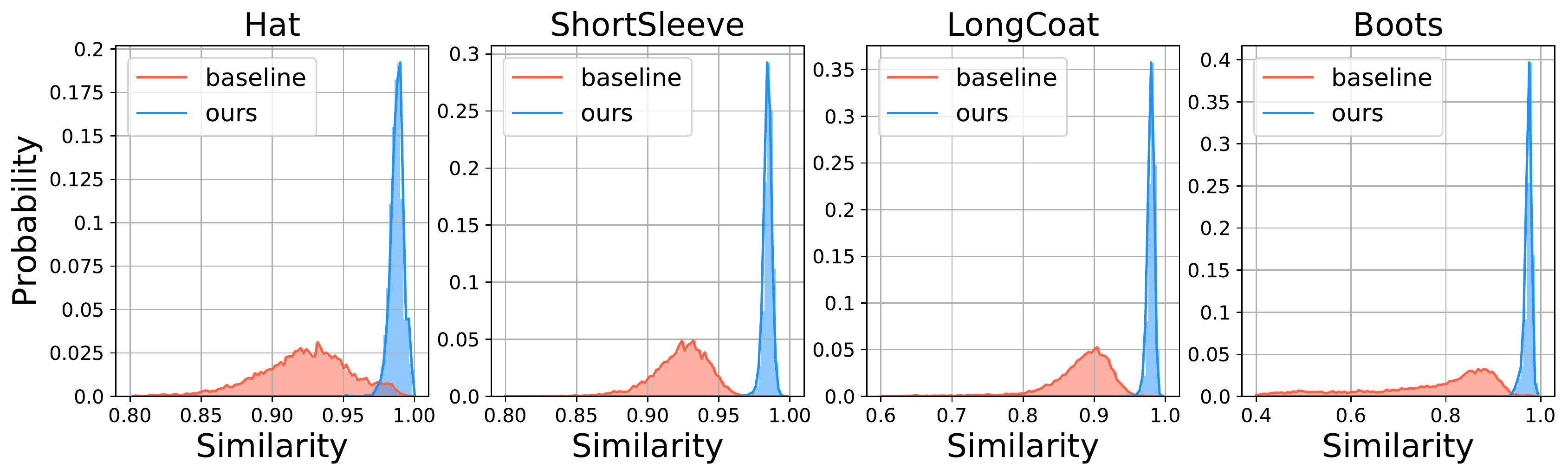}}
		
	\caption{{\bf Illustration of the effect of the SPAC and SEMC module.} We take the ``hat", ``short sleeve", ``long-coat", and ``boots" attributes as examples to show (a) the spatial attention regions, (b) similarity distribution of spatial attention regions, and (c) similarity distribution of semantic features between different images of the same attribute. Compared with the baseline method, most similarities of the proposed method are concentrated near 1, which proves the consistency of spatial attention regions and semantic features for each attribute.}
	\label{fig:effect}
	\vspace{-1.5em}
\end{figure}

To validate the effectiveness of the proposed SPAC module and SEMC module, we visualize the spatial attention regions and the similarity distributions of spatial and semantic features                    in Figure \ref{fig:effect}. The similarities are computed between each pair of images of the same attribute on PA100K. The higher the similarity, the more consistent the attention regions and semantic features of the two images with the same attribute. Compared to the baseline method, as shown in Figure \ref{fig:effect}(b) and Figure \ref{fig:effect}(c), we observe that plenty of similarities concentrate on 1, making the probability curve rise rapidly near 1. The same phenomenon can also be observed in other attributes of the PA100K, RAP, and PETA as shown in supplementary material.

\subsection{Hyperparameter Evaluation}

There are mainly three key hyperparameters in our method, which are confidence threshold $\tau$, initial epoch $i_{e}$, and momentum coefficient $\alpha$. We set $\tau = 0.9$, $i_{e}=4$, $\alpha=0.9$ if not specially specified. To fully demonstrate the effect of hyperparameters, the following experiments are all conducted on the largest pedestrian attribute dataset PA100K.

\begin{table}[h]
\centering
\caption{ {\bf Experiments on the confidence threshold $\tau$}.} 
\resizebox{\columnwidth}{!}{
\begin{tabular}{c|c|c|c|c|c}
\toprule
	Confidence Threshold & mA & Accu & Prec & Recall & F1 \\\midrule \midrule
	$\tau = 0 \ \  \ $ & 79.63 & 78.61 & \bf{86.89} & 87.19 & 86.63 \\
	$\tau = 0.3$ & 80.08 & 78.21 & 86.65 & 86.88 & 86.35 \\
	$\tau = 0.5$ & 80.90 & 78.40 & 86.49 & 87.36 & 86.51 \\
	$\tau = 0.7$ & 80.79 & 78.20 & 86.37 & 87.14 & 86.35 \\
	$\tau = 0.9$ & \bf{81.87} & \bf{78.86} & 85.98 & \bf{89.10} & \bf{86.87} \\
\bottomrule
\end{tabular}}
\label{tab:hyper_tau}
\vspace{-1.5em}
\end{table}

Confidence threshold $\tau$ is used in Equation \ref{eq:tau_a} and Equation \ref{eq:tau_v} to select reliable spatial attention feature maps $\bm{A}^{q}$ and semantic feature vectors $\bm{V}^{q}$, which are aggregated to spatial memory $\bm{M}^{spa}$ and semantic memory $\bm{M}^{sem}$. As shown in Table \ref{tab:hyper_tau}, with the increase of the confidence threshold $\tau$, there is an obvious performance improvement in mA from 79.63 to 81.28. It is easy to infer that higher confidence threshold $\tau$ can select more precise spatial attention feature maps and more discriminative semantic feature vectors. Little performance fluctuation in the other four metrics shows the robustness of the threshold $\tau$.

\begin{table}[h]
\centering
\caption{{\bf Experiments on the momentum coefficient $\alpha$}.} 
\resizebox{\columnwidth}{!}{
\begin{tabular}{c|c|c|c|c|c}
\toprule
	Momentum Coefficient & mA & Accu & Prec & Recall & F1 \\\midrule \midrule
	$\alpha = 0.1$ & 80.89 & 78.23 & 86.45 & 87.12 & 86.37 \\
	$\alpha = 0.3$ & 81.04 & 78.19 & 86.37 & 87.05 & 86.32 \\
	$\alpha = 0.5$ & 81.10 & 78.17 & 86.31 & 87.19 & 86.33 \\
	$\alpha = 0.7$ & 81.15 & 78.34 & \bf{86.46} & 87.35 & 86.49 \\
	$\alpha = 0.9$ & \bf{81.87} & \bf{78.86} & 85.98 & \bf{89.10} & \bf{86.87} \\
	\bottomrule
\end{tabular}}
\label{tab:hyper_alpha}
\vspace{-1em}
\end{table}

Momentum coefficient $\alpha$ is adopted in Equation \ref{eq:alpha_a} and Equation \ref{eq:alpha_v} to determine the degree of integration of historical features and current batch features. The larger $\alpha$ is, the fewer historical features are retained. As shown in Table \ref{tab:hyper_alpha}, more historical features can bring a few performance improvements.

\vspace{-0.5em}
\section{Conclusion} \label{conclusion}

This paper proposes the consistency framework for pedestrian attribute recognition, which makes full use of the inter-image relation of the same attribute and tackles the spatial attention region deviation problem. Specifically, we propose the SPAC module to pay attention to specific attribute-related spatial regions. We also propose the SEMC module to extract intrinsic and discriminative semantic features for each attribute. Moreover, we implement two variants of our method to demonstrate the efficacy of consistency regularizations. The ablation experiments show that two consistency modules can both bring performance improvements. Our proposed method achieves outstanding performance consistently on the PA100K, RAP, and PETA.

\vspace{-0.5em}
\section{Acknowledgments} \label{ack}

This work was supported in part by the National Natural Science Foundation of China (Grant No.61721004 and Grant No.61876181), the Projects of Chinese Academy of Science (Grant QYZDB-SSW-JSC006), the Strategic Priority Research Program of Chinese Academy of Sciences (Grant No. XDA27000000), and the Youth Innovation Promotion Association CAS.

\clearpage

{\small
\bibliographystyle{iccv}
\bibliography{iccv}

\begin{thebibliography}{10}\itemsep=-1pt

\bibitem{deng2014pedestrian}
Yubin Deng, Ping Luo, Chen~Change Loy, and Xiaoou Tang.
\newblock Pedestrian attribute recognition at far distance.
\newblock In {\em Proceedings of the 22nd ACM international conference on
  Multimedia}, pages 789--792, 2014.

\bibitem{guo2019visual}
Hao Guo, Kang Zheng, Xiaochuan Fan, Hongkai Yu, and Song Wang.
\newblock Visual attention consistency under image transforms for multi-label
  image classification.
\newblock In {\em Proceedings of the IEEE Conference on Computer Vision and
  Pattern Recognition}, pages 729--739, 2019.

\bibitem{he2016deep}
Kaiming He, Xiangyu Zhang, Shaoqing Ren, and Jian Sun.
\newblock Deep residual learning for image recognition.
\newblock In {\em Proceedings of the IEEE Conference on Computer Vision and
  Pattern Recognition}, pages 770--778, 2016.

\bibitem{hochreiter1997long}
Sepp Hochreiter and J{\"u}rgen Schmidhuber.
\newblock Long short-term memory.
\newblock {\em Neural computation}, 9(8):1735--1780, 1997.

\bibitem{TCLNet}
Ruibing Hou, Hong Chang, Bingpeng Ma, Shiguang Shan, and Xilin Chen.
\newblock Temporal complementary learning for video person re-identification.
\newblock In {\em Proceedings of the European Conference on Computer Vision},
  2020.

\bibitem{hu2018squeeze}
Jie Hu, Li Shen, and Gang Sun.
\newblock Squeeze-and-excitation networks.
\newblock In {\em Proceedings of the IEEE Conference on Computer Vision and
  Pattern Recognition}, pages 7132--7141, 2018.

\bibitem{huang2020improve}
Houjing Huang, Wenjie Yang, Jinbin Lin, Guan Huang, Jiamiao Xu, Guoli Wang,
  Xiaotang Chen, and Kaiqi Huang.
\newblock Improve person re-identification with part awareness learning.
\newblock {\em IEEE Transactions on Image Processing}, 29:7468--7481, 2020.

\bibitem{jaderberg2015spatial}
Max Jaderberg, Karen Simonyan, Andrew Zisserman, et~al.
\newblock Spatial transformer networks.
\newblock In {\em Advances in Neural Information Processing Systems}, pages
  2017--2025, 2015.

\bibitem{jia2020rethinking}
Jian Jia, Houjing Huang, Wenjie Yang, Xiaotang Chen, and Kaiqi Huang.
\newblock Rethinking of pedestrian attribute recognition: Realistic datasets
  with efficient method.
\newblock {\em arXiv preprint arXiv:2005.11909}, 2020.

\bibitem{li2015deepmar}
Dangwei Li, Xiaotang Chen, and Kaiqi Huang.
\newblock Multi-attribute learning for pedestrian attribute recognition in
  surveillance scenarios.
\newblock In {\em 2015 3rd IAPR Asian Conference on Pattern Recognition}, pages
  111--115. IEEE, 2015.

\bibitem{li2018pose}
Dangwei Li, Xiaotang Chen, Zhang Zhang, and Kaiqi Huang.
\newblock Pose guided deep model for pedestrian attribute recognition in
  surveillance scenarios.
\newblock In {\em 2018 IEEE International Conference on Multimedia and Expo},
  pages 1--6. IEEE, 2018.

\bibitem{li2018richly}
Dangwei Li, Zhang Zhang, Xiaotang Chen, and Kaiqi Huang.
\newblock A richly annotated pedestrian dataset for person retrieval in real
  surveillance scenarios.
\newblock {\em IEEE transactions on image processing}, 28(4):1575--1590, 2018.

\bibitem{li2016richly}
Dangwei Li, Zhang Zhang, Xiaotang Chen, Haibin Ling, and Kaiqi Huang.
\newblock A richly annotated dataset for pedestrian attribute recognition.
\newblock {\em arXiv preprint arXiv:1603.07054}, 2016.

\bibitem{lin2017feature}
Tsung-Yi Lin, Piotr Doll{\'a}r, Ross Girshick, Kaiming He, Bharath Hariharan,
  and Serge Belongie.
\newblock Feature pyramid networks for object detection.
\newblock In {\em Proceedings of the IEEE Conference on Computer Vision and
  Pattern Recognition}, pages 2117--2125, 2017.

\bibitem{liu2018localization}
Pengze Liu, Xihui Liu, Junjie Yan, and Jing Shao.
\newblock Localization guided learning for pedestrian attribute recognition.
\newblock {\em arXiv preprint arXiv:1808.09102}, 2018.

\bibitem{liu2017hydraplus}
Xihui Liu, Haiyu Zhao, Maoqing Tian, Lu Sheng, Jing Shao, Shuai Yi, Junjie Yan,
  and Xiaogang Wang.
\newblock Hydraplus-net: Attentive deep features for pedestrian analysis.
\newblock In {\em Proceedings of the IEEE International Conference on Computer
  Vision}, pages 350--359, 2017.

\bibitem{sarafianos2018deep}
Nikolaos Sarafianos, Xiang Xu, and Ioannis~A Kakadiaris.
\newblock Deep imbalanced attribute classification using visual attention
  aggregation.
\newblock In {\em Proceedings of the European Conference on Computer Vision},
  pages 680--697, 2018.

\bibitem{shao2015deeply}
Jing Shao, Kai Kang, Chen Change~Loy, and Xiaogang Wang.
\newblock Deeply learned attributes for crowded scene understanding.
\newblock In {\em Proceedings of the IEEE Conference on Computer Vision and
  Pattern Recognition}, pages 4657--4666, 2015.

\bibitem{tang2019Improving}
Chufeng Tang, Lu Sheng, Zhaoxiang Zhang, and Xiaolin Hu.
\newblock Improving pedestrian attribute recognition with weakly-supervised
  multi-scale attribute-specific localization.
\newblock In {\em Proceedings of the IEEE International Conference on Computer
  Vision}, pages 4997--5006, 2019.

\bibitem{wang2017attribute}
Jingya Wang, Xiatian Zhu, Shaogang Gong, and Wei Li.
\newblock Attribute recognition by joint recurrent learning of context and
  correlation.
\newblock In {\em Proceedings of the IEEE International Conference on Computer
  Vision}, pages 531--540, 2017.

\bibitem{wang2019parsurvey}
Xiao Wang, Shaofei Zheng, Rui Yang, Bin Luo, and Jin Tang.
\newblock Pedestrian attribute recognition: A survey.
\newblock {\em arXiv preprint arXiv:1901.07474}, 2019.

\bibitem{yang2020hierarchical}
Jie Yang, Jiarou Fan, Yiru Wang, Yige Wang, Weihao Gan, Lin Liu, and Wei Wu.
\newblock Hierarchical feature embedding for attribute recognition.
\newblock In {\em Proceedings of the IEEE Conference on Computer Vision and
  Pattern Recognition}, pages 13055--13064, 2020.

\bibitem{yang2019towards}
Wenjie Yang, Houjing Huang, Zhang Zhang, Xiaotang Chen, Kaiqi Huang, and Shu
  Zhang.
\newblock Towards rich feature discovery with class activation maps
  augmentation for person re-identification.
\newblock In {\em Proceedings of the IEEE Conference on Computer Vision and
  Pattern Recognition}, pages 1389--1398, 2019.

\bibitem{yu2016weakly}
Kai Yu, Biao Leng, Zhang Zhang, Dangwei Li, and Kaiqi Huang.
\newblock Weakly-supervised learning of mid-level features for pedestrian
  attribute recognition and localization.
\newblock {\em arXiv preprint arXiv:1611.05603}, 2016.

\bibitem{zhou2016learning}
Bolei Zhou, Aditya Khosla, Agata Lapedriza, Aude Oliva, and Antonio Torralba.
\newblock Learning deep features for discriminative localization.
\newblock In {\em Proceedings of the IEEE Conference on Computer Vision and
  Pattern Recognition}, pages 2921--2929, 2016.

\bibitem{zhu2013pedestrian}
Jianqing Zhu, Shengcai Liao, Zhen Lei, Dong Yi, and Stan~Z Li.
\newblock Pedestrian attribute classification in surveillance: Database and
  evaluation.
\newblock In {\em 2013 IEEE International Conference on Computer Vision
  Workshops}, pages 331--338. IEEE, 2013.

\bibitem{zitnick2014edge}
C~Lawrence Zitnick and Piotr Doll{\'a}r.
\newblock Edge boxes: Locating object proposals from edges.
\newblock In {\em Proceedings of the European Conference on Computer Vision},
  pages 391--405, 2014.

\end{thebibliography}
}

\end{document}